\documentclass[conference]{IEEEtran}
\IEEEoverridecommandlockouts
\usepackage{ragged2e} 
\usepackage{booktabs,makecell, multirow, tabularx}
\usepackage{cite}
\usepackage{amsmath,amssymb,amsfonts}
\usepackage{algorithmic}
\usepackage{graphicx}
\usepackage{textcomp}
\usepackage{xcolor}
\usepackage[numbers,sort&compress]{natbib}
\def\BibTeX{{\rm B\kern-.05em{\sc i\kern-.025em b}\kern-.08em
    T\kern-.1667em\lower.7ex\hbox{E}\kern-.125emX}}
\begin{document}

\title{LKASeg:Remote-Sensing Image Semantic Segmentation with Large Kernel Attention and Full-Scale Skip Connections\\
\thanks{This work was supported in part by the National Natural Science Foundation of China under Grant 62271160 and 62176068, in part by the Natural Science Foundation of Heilongjiang Province of China under Grant LH2021F011, in part by the Fundamental Research Funds for the Central Universities of China under Grant 3072024LJ0803, in part by the Natural Science Foundation of Guangdong Province of China under Grant 2022A1515011527.}
}

\author{\IEEEauthorblockN{1\textsuperscript{st}Xuezhi Xiang}
	\IEEEauthorblockA{\textit{Harbin Engineering University} \\
		Harbin, China \\
		xiangxuezhi@hrbeu.edu.cn}
	\and

	\IEEEauthorblockN{2\textsuperscript{nd} Yibo Ning}
	\IEEEauthorblockA{\textit{Harbin Engineering University} \\
		Harbin, China \\
		13878109862@163.com}
	\and
	
	\IEEEauthorblockN{3\textsuperscript{rd} Lei Zhang}
	\IEEEauthorblockA{\textit{Guangdong University of Petrochemical Technology} \\
		Maoming, China \\
		zhanglei@gdupt.edu.cn}
	\and

	\and
	\IEEEauthorblockN{4\textsuperscript{th} Denis Ombati}
	\IEEEauthorblockA{\textit{Harbin Engineering University} \\
		Harbin, China \\
		deniso2009@gmail.com}
	
	\and
	
	\IEEEauthorblockN{5\textsuperscript{th} Himaloy Himu}
	\IEEEauthorblockA{\textit{Harbin Engineering University} \\
		Harbin, China \\
		himaloy@hrbeu.edu.cn}
	
	\and
	
	\IEEEauthorblockN{6\textsuperscript{th} Xiantong Zhen}
	\IEEEauthorblockA{\textit{Guangdong University of Petrochemical Technology} \\
		Maoming, China \\
		zhenxt@gmail.com}

}

\maketitle

\begin{abstract}
Semantic segmentation of remote sensing images is a fundamental task in geospatial research. However, widely used Convolutional Neural Networks (CNNs) and Transformers have notable drawbacks: CNNs may be limited by insufficient remote sensing modeling capability, while Transformers face challenges due to computational complexity. In this paper, we propose a remote-sensing image semantic segmentation network named LKASeg, which combines Large Kernel Attention(LSKA) and Full-Scale Skip Connections(FSC). Specifically, we propose a decoder based on Large Kernel Attention (LKA), which extract global features while avoiding the computational overhead of self-attention and providing channel adaptability. To achieve full-scale feature learning and fusion, we apply Full-Scale Skip Connections (FSC) between the encoder and decoder. We conducted experiments by combining the LKA-based decoder with FSC. On the ISPRS Vaihingen dataset, the mF1 and mIoU scores achieved 90.33\% and 82.77\%.
\end{abstract}


\section{Introduction}
With the widespread application of deep neural networks in computer vision, significant advancements have been made in semantic segmentation of natural images, achieving remarkable results. Notably, Long et al. \cite{7298965} introduced Fully Convolutional Networks (FCN), which enable end-to-end dense pixel prediction. However, the overly simplistic upsampling in FCN leads to less precise segmentation results. Subsequently, Ronneberger et al. \cite{Ronneberger2015UNetCN} proposed UNet, an encoder-decoder network with a symmetric structure. UNet improves segmentation accuracy by using skip connections to compensate for the loss of feature information during downsampling. However, UNet merely concatenates low-level features generated by the encoder with high-level features from the decoder, without further refining these features, which limits the network's ability to differentiate between low-level and high-level features.

Methods based on convolution, such as FCN and UNet, lack the ability to capture global information. In high-resolution remote-sensing image, where complex objects frequently appear, the absence of global semantic information makes precise segmentation challenging. To capture global semantic information, one solution is to incorporate transformer-based self-attention mechanisms. Existing transformer-based semantic segmentation methods for remote-sensing image can be broadly categorized into two types. The first type consists of encoder-decoder architectures composed entirely of transformers, such as SegFormer \cite{Xie2021SegFormerSA}. The second type is hybrid encoder-decoder architectures combining CNNs and transformers. For example, Wang et al. \cite{Wang2021TransformerMC} proposed a Bidirectional Attention Network (BANet) with dependency and texture pathways, where the dependency pathway uses transformers to capture long-range dependencies, and the texture pathway uses CNNs to extract local texture features. Wang et al. \cite{Wang2021UNetFormerAU} introduced UNetFormer, a transformer-like architecture that innovatively employs a hybrid design, combining a CNN-based encoder with a decoder based on Global-Local Transformer Blocks (GLTB). Another approach involves using large kernel convolutions \cite{Woo2018CBAMCB} to build correlations and generate attention maps to capture global information.

However, large kernel convolutions and self-attention still have their drawbacks. Large kernel convolutions incur significant computational overhead and parameter costs, and the static nature of convolutional weights lacks adaptability. Guo et al. \cite{Guo2022VisualAN} noted that self-attention was originally designed for one-dimensional NLP tasks, treating two-dimensional images as one-dimensional sequences, which disrupts the critical two-dimensional structure of images. Additionally, the quadratic computation and memory overhead of self-attention make it challenging to handle high-resolution images. Furthermore, self-attention is a mechanism that only considers spatial adaptability while neglecting channel-wise adaptability. We address these issues by introducing Large Kernel Attention (LKA), which leverages the advantages of both self-attention and convolution. LKA combines the benefits of convolution and self-attention, including local structural information, long-range dependencies, and adaptability, while also overcoming the limitation of self-attention's lack of channel-wise adaptability.

In practical applications, remote sensing images often contain numerous objects, with similar objects arranged densely, differing greatly in size, color, and texture. The issue of scale variation in remote sensing images is thus a significant concern. Spatial detail information is essential for accurate semantic segmentation. Convolutional Neural Networks (CNNs) can easily lose spatial and boundary information during the feature extraction phase of semantic segmentation. Research on segmentation shows that feature maps at different scales provide different types of information. Huang et al. \cite{9053405} noted that lower-level detailed feature maps capture rich spatial information, while higher-level semantic feature maps contain positional information. TransUNet adopts hybrid visual transformers \cite{Chen2021TransUNetTM} as encoders to enhance feature extraction and achieves state-of-the-art results in medical image segmentation. Li et al. \cite{Wang2021UNetFormerAU} proposed UNetFormer, which uses skip connections of the same scale between the encoder and decoder to mitigate the loss of spatial details. Wu et al. \cite{10247595} introduced CMTFNet, which employs multi-scale transformers for multi-scale feature learning and fusion. However, these methods do not directly integrate feature maps of different scales from a full-size perspective. In this paper, we use full-size skip connections to address the issues of scale variation and spatial information loss in remote sensing image segmentation tasks. The contributions of this work can be summarized as follows:

1. We propose a remote-sensing image semantic segmentation network named LKASeg. We introduce a decoder based on Large Kernel Attention (LKA), which extract global features and provide channel adaptability while avoiding the computational overhead of self-attention.

2. We apply Full-Scale Skip Connections (FSC) between the encoder and decoder in LKASeg to achieve full-scale feature learning and fusion, addressing issues of scale variation and spatial information loss in remote sensing image semantic segmentation tasks.

3. We conducted experiments by combining the LKA-based decoder with FSC. On the ISPRS Vaihingen dataset, the mF1 and mIoU scores achieved 90.33\% and 82.77\%.

\section{METHODOLOGY}
\subsection{Overall Architecture}
The overall architecture of LKASeg is shown in
Fig. 1. LKASeg consists of three main components: an encoder based on ResNet-18, a decoder based on Large Kernel Attention (LKA), and full-size skip connections (FSC). We chose the pretrained ResNet-18 as the encoder to extract multiscale semantic features at very low computational costs. ResNet-18 consists of four stages of Resblocks, each downsampling the feature maps by a factor of 2. We employed the same Feature Refinement Head (FRH) design as the baseline. Additionally, we aggregate the semantic features generated by each layer of Resblocks with those generated by the LKA-based decoder below using a weighted sum operation. This operation selectively weights the contributions of both features to segmentation accuracy, thereby learning more generalized fused features. The weighted sum operation is as:
\begin{equation}
	F_f = \alpha \times F_R + (1 - \alpha) \times F_L ,
\end{equation}
where $F_f$ is the fused features, $F_R$ is the features generated by Resblocks, and $F_L$ is the feature generated by the decoder blocks composed of LKA. A detailed description of the decoder, which is based LKA and FSC will show in the following sections.

\subsection{LKA-based decoder}\label{AA}
LKA is derived by decomposing a large kernel convolution.
\begin{figure}[t!]
	\centering
	\centerline{\includegraphics[scale=0.37]{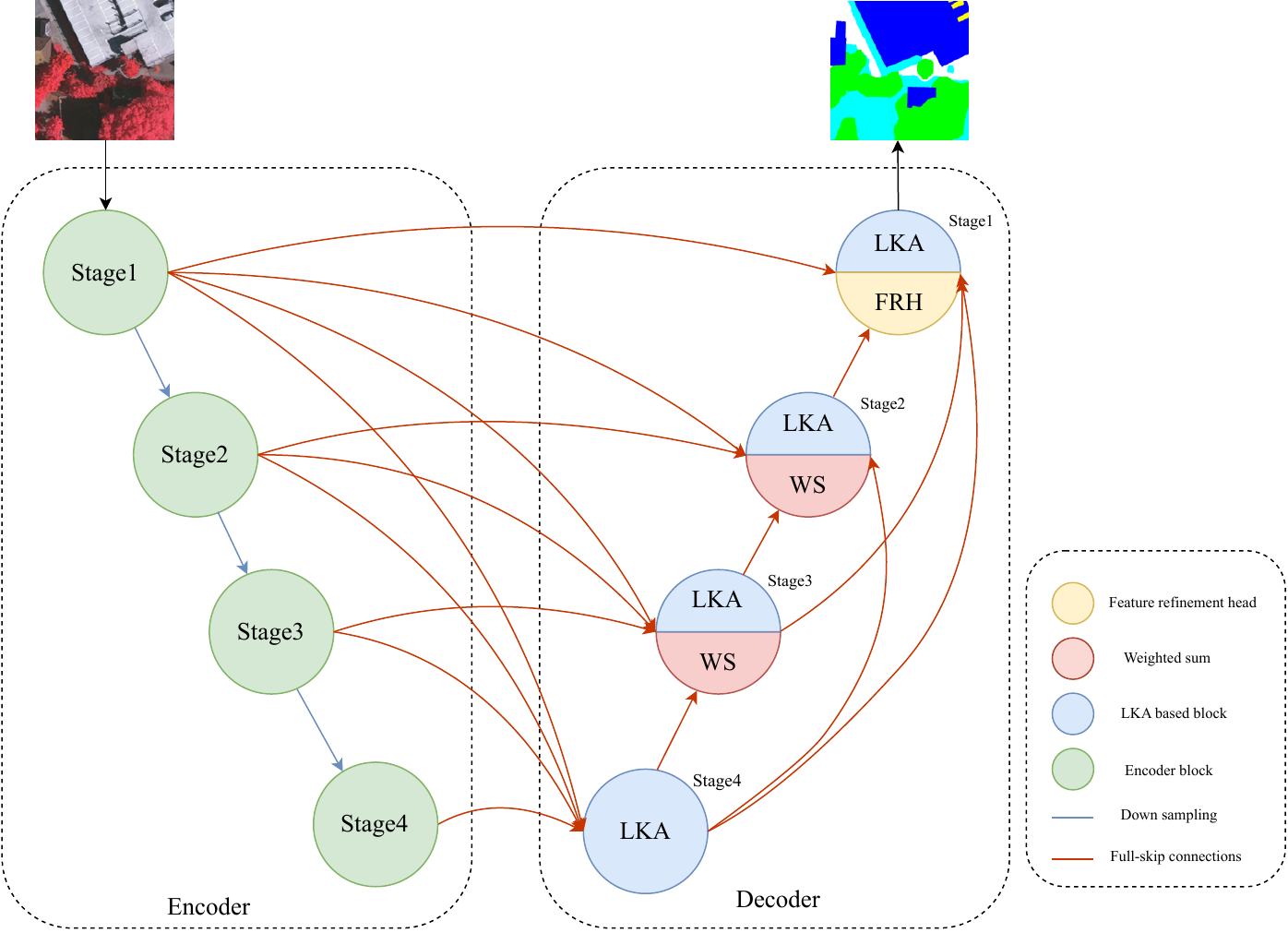}}
	\caption{Illustration of the LKASeg.}
	\label{fig}
\end{figure}
As shown in Fig. 2, a $K \times K $ convolution can be decomposed into three parts: depthwise local convolution, depthwise dilated convolution with expansion factor $d $, and channel convolution $1\times1$ convolution. Specifically, a $K \times K $ convolution can be decomposed into a $\frac{K}{d} \times \frac{K}{d} $ depthwise expansion convolution, a $(2d$ -$1) \times (2d$ -$1) $ depthwise convolution, and a $1\times1$ convolution. Through this decomposition, we can capture long-range dependencies with slight computational cost and parameters, and channel adaptability is achieved through $1\times1$ convolution. After obtaining long-range dependencies, we estimate the importance of each point and generate attention maps. The operations of LKA module can be expressed as:
\begin{equation}
	Attention = Conv1\times1(DW-D-Conv(DW-Conv(F))) ,
\end{equation} 
\begin{equation}
	Output = Attention \otimes F_i ,
\end{equation}
here, $F_i \in C\times H\times W $ represents the input features, $Conv1\times1$ represents $1\times1$ convolution, $DW-D-Conv$ represents  depth-wise dilation convolution, $DW-Conv$ represents depth-wise convolution, $\otimes $ denotes element-wise product, and $Attention \in  C\times H\times W $ denotes the attention map. The value in attention map indicates the importance of each feature.  LKA combines the benefits of convolution and self-attention, considering local context information, large receptive fields, linear complexity, and dynamic processes. Moreover, LKA achieves adaptability not only in the spatial dimension but also in the channel dimension.

\subsection{Full-Scale Skip Connections}
\begin{figure}[htbp]
	\centerline{\includegraphics[scale=0.8]{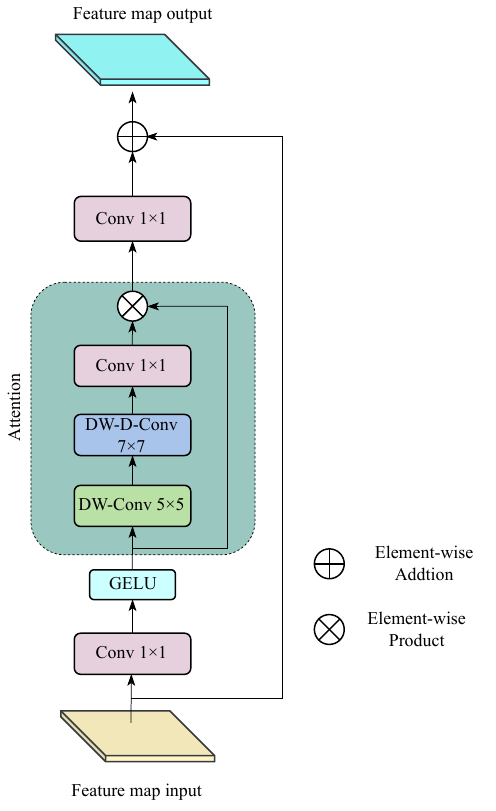}}
	\caption{Illustration of the LKA.}
	\label{fig}
\end{figure}
To fully utilize feature information at all scales, we use FSC between features output by the encoder and those output by the decoder units. As shown in Fig. 1. Adjacent decoder units are connected via FSC pathways. First, features at four different scales are converted to a common scale for fusion. In our method, feature maps generated at each encoder stage are combined with the corresponding feature maps from the decoder through a weighted sum, and then processed through the LKA module, thus aggregating features from both the encoder and decoder. We use max pooling operations to reduce the dimensionality of the features and bilinear interpolation to scale up the feature dimensions. We employ linear layers to unify the number of feature channels, which helps in reducing redundant information and model parameters. Finally, four feature maps, harmonized in scale and channel count, are concatenated along the channel direction and then fed into the decoding unit for further feature fusion. Our FSC method leverages full-scale feature information, thoroughly integrating relationships between different levels of features, and preserving more spatial details.

\section{EXPERIMENTS AND DISCUSSION}
\subsection{Datasets}

ISPRS Vaihingen:
The ISPRS Vaihingen dataset consists of 16 very high-resolution true orthophotos, averaging 2500 × 2000 pixels each. Each orthophoto includes three channels: near-infrared, red, and green (NIRG), with a ground sampling distance of 9 centimeters. The dataset comprises five foreground classes: impervious surfaces, buildings, low vegetation, trees, cars, and one background class (clutter). The 16 orthophotos are divided into a training set of 12 patches and a test set of 4 patches. The training set includes orthophotos indexed as 1, 3, 23, 26, 7, 11, 13, 28, 17, 32, 34, and 37, while the test set includes orthophotos indexed as 5, 21, 15, and 30.

\subsection{Implementation Details}
We use ResNet18-based UNetformer \cite{Wang2021UNetFormerAU} as the baseline model. The proposed LSKASeg is benchmarked against several prominent methods, including ABCNet \cite{Li2021ABCNetAB}, TransUNet \cite{Chen2021TransUNetTM}, UNetformer \cite{Wang2021UNetFormerAU}, CMTFNet \cite{10247595}, BANet \cite{Wang2021TransformerMC}, and MARes-Unet \cite{9378788}. Experiments are conducted using PyTorch on a single NVIDIA GeForce RTX 4090 GPU with 24GB RAM. All models are optimized and trained using the Stochastic Gradient Descent (SGD) algorithm with a learning rate of 0.01, momentum of 0.9, weight decay of 0.0005, and a batch size of 10. The total number of epochs is set to 50, with testing conducted after each epoch. To quantitatively evaluate the proposed method, two widely used metrics are recorded: mean F1 score (mF1) and mean Intersection over Union (mIoU).

\begin{table*}[]
	\caption{EXPERIMENTAL RESULTS ON THE ISPRS VAIHINGEN DATASET. WE PRESENT THE OA OF FIVE FOREGROUND CLASSES AND THREE OVERALL
		PERFORMANCE METRICS. THE ACCURACY OF EACH CATEGORY IS PRESENTED IN THE F1/IOU FORM. BOLD VALUES ARE THE BEST.}
	\centering
	\begin{tabular}{ccccccccc}
		\hline
		\textbf{Method}     & \textbf{Backbone}  & \textbf{impervious surface} & \textbf{building}    & \textbf{low vegetation} & \textbf{tree}        & \textbf{car}         & \textbf{mF1}   & \textbf{mIoU}  \\ \hline
		MARes-Unet\cite{9378788} & ResNet-34 & 91.91/ 85.02       & 96.04/92.37 & 80.03/66.71    & 90.44/82.54 & 88.83/79.90 & 89.45 & 81.31 \\
		BANet\cite{Wang2021TransformerMC}      & ResT-Lite & 91.89/84.99        & 96.21/92.69 & 80.26/67.02    & 90.91/83.33 & 86.11/75.60 & 89.07 & 80.73 \\
		ABCNet\cite{Li2021ABCNetAB}     & ResNet-18 & 91.60/84.51        & 96.11/92.50 & 77.24/62.91    & 89.52/81.03 & 74.71/59.63 & 85.84 & 76.12 \\
		TransUNet\cite{Chen2021TransUNetTM}  & R50-ViT-B & 90.72/83.02        & 94.43/89.45 & 79.42/65.87    & 90.63/82.87 & 83.97/72.36 & 87.84 & 78.72 \\
		UNetformer\cite{Wang2021UNetFormerAU} & ResNet-18 & 91.97/85.13        & 96.15/92.58 & 79.99/66.66    & 90.88/83.29 & 90.27/82.27 & 89.85 & 81.98 \\
		CMTFNet\cite{10247595}    & ResNet-50 & 91.70/84.67        & 95.88/92.07 & 81.00/68.07    & 91.01/83.50 & 90.21/82.17 & 89.96 & 82.10 \\ \hline
		LKASeg(Ours)      & ResNet-18 & \textbf{92.52/86.08}        & \textbf{96.71/93.63} & 80.82/67.81    & \textbf{91.08/83.61} & \textbf{90.53/82.70} & \textbf{90.33} & \textbf{82.77} \\ \hline
	\end{tabular}
\end{table*}

\subsection{Performance Comparison}

\begin{figure}[!t]
	\centering
	\includegraphics[scale=0.4]{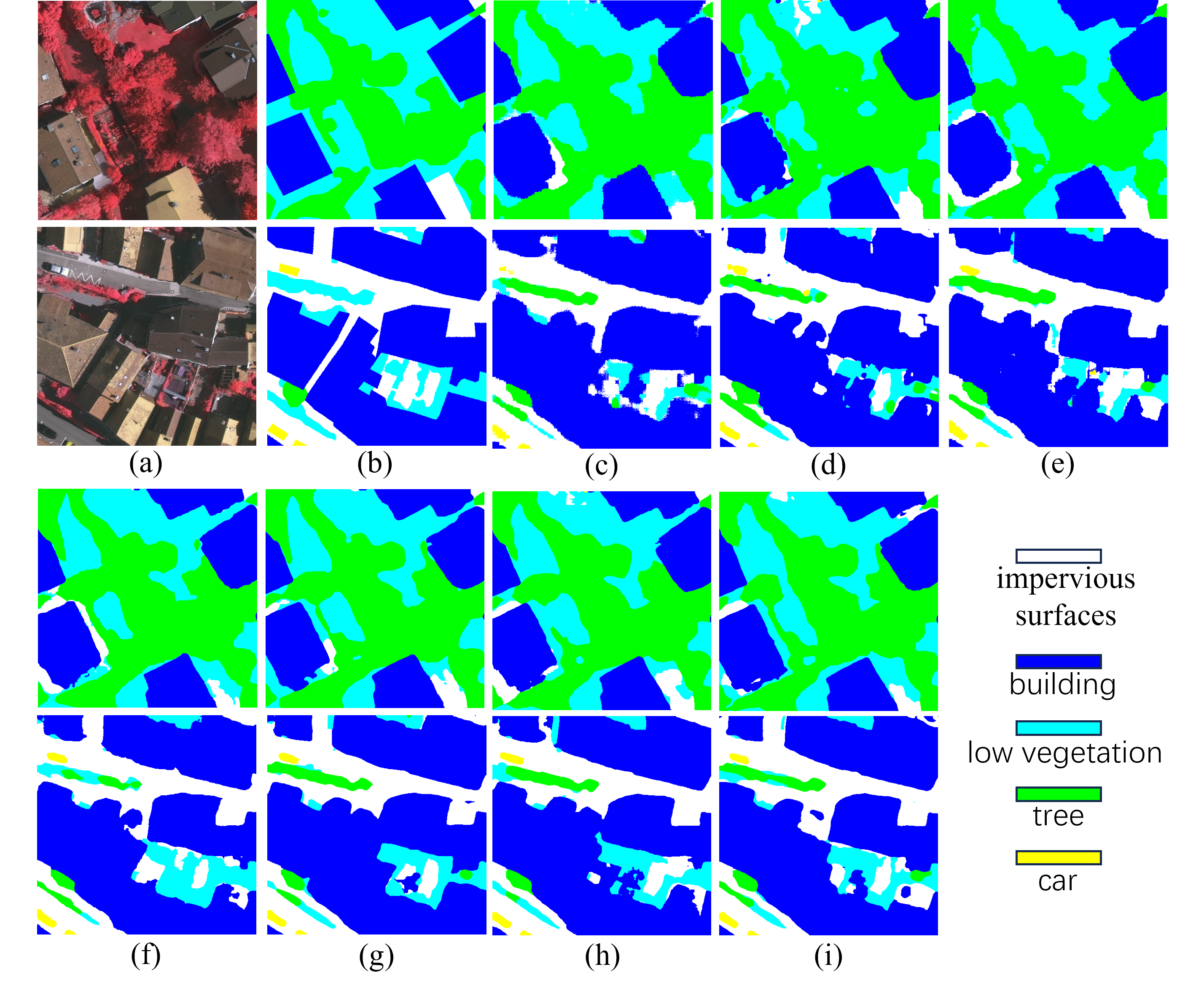}
	\caption{Qualitative performance comparisons on the ISPRS Vaihaigen with
		the size of 512 × 512. (a) NIRRG images, (b) Ground truth, (c) ABCNet, (d) TransUNet, (e) BANet, (f) MARes-Unet, (g) UNetformer, (h) CMTFNet and (i) LKASeg.}
	\label{fig_1}
\end{figure}

Performance comparison on the Vaihingen dataset: As shown in Table 1, the proposed LKASeg shows significant improvements in mF1 and mIoU compared to the baseline UNetformer. This confirms the effectiveness of the LKA-based decoder structure and full-scale skip connections. Compared to several models included in the comparison, LKASeg performs excellently across most categories. Notably, LKASeg's F1 and IoU increased by 0.52\% and 1.01\% for the building category compared to BANet. The IoU for impervious surface increased by 0.60\% and for car by 0.29\% compared to the baseline UNetformer. Overall, LKASeg achieved mF1 and mIoU scores of 90.33\% and 82.77\%, respectively, mF1 and mIoU increased by 0.53\% and 0.96\% to the baseline UNetformer. These improvements are mainly attributed to the global semantic information from LKA-based decoder and full-scale features from full-scale skip connections. Fig. 3 shows visual examples of results obtained by all eight methods. Clearly, LKASeg can more accurately segment objects, with smoother boundaries and fewer noise points.

\subsection{Ablation Study}
In this section, we evaluate the model performance under different configurations and compare our method with the baseline \cite{Wang2021UNetFormerAU}. First, without the LKA module and FSC structure, the model's ability to obtain global information and full-scale information is insufficient, leading to decreased performance. Next, we evaluated networks using only the LKA module, only the FSC structure, and using both LKA and FSC structures. All four configurations outperformed the baseline network. This is because the LKA modules learn long-range dependencies between pixels without disrupting the 2D structure of the context features, maintaining channel adaptivity and significantly improving semantic segmentation results for remote sensing images. As shown in the table 2, replacing the baseline transformer-based decoder with LKA-based decoder not only improves mF1 and mIoU but also reduces the number of parameters and computational cost, demonstrating the effectiveness of our structure and its simplified computational complexity. The FSC effectively handles large variations in shape and size through dense encoder-decoder connections. Although it results in a slight increase in the number of parameters and computational cost, it improves segmentation accuracy. Compared to the baseline, our method's mF1 and mIoU on the Vaihingen dataset increased by 0.53\% and 0.96\%, respectively.

\begin{table}[]
	\caption{ABLATION STUDIES}
	\begin{tabular}{lcccc}
		\hline
		\textbf{Method}            & \textbf{mF1}   & \textbf{mIoU}  & \textbf{Parameters(M)}   & \textbf{FLOPs(G)} \\ \hline
		Baseline          & 89.85 & 81.98 & 11.69 M  & 5.88 G   \\
		Baseline+LKA      & 90.17 & 82.49 & 11.55 M  & 5.76 G   \\
		Baseline+FSC      & 90.16 & 82.45 & 16.52 M & 21.62 G   \\
		Baseline+LKA+FSC  & \textbf{90.33} & \textbf{82.77} & 15.78 M & 19.68 G   \\ \hline
	\end{tabular}
\end{table}

\section{CONCLUSION}

This paper proposes a new semantic segmentation network, LKASeg, which features two decoders based on Large Kernel Attention (LKA), respectively. The LKA-based decoder extract global features and offer channel adaptability while avoiding the computational overhead of self-attention. To achieve full-scale feature learning and fusion, we apply Full-Scale Skip Connections (FSC) between the encoder and decoder. We conducted experiments combining LKA-based decoder with FSC. Evaluations on a remote sensing datasets demonstrate that LKASeg outperforms existing CNN and Transformer-based semantic segmentation methods. Specifically, on the ISPRS Vaihingen dataset, the mF1 and mIoU scores increased by 0.53\% and 0.96\%, respectively, compared to the baseline. These results highlight the potential of our method in the field of remote sensing semantic segmentation.

\newpage

\bibliographystyle{IEEEtran}
\bibliography{refs}

\end{document}